\documentclass{article}

\usepackage[preprint, nonatbib]{nbvae}

\usepackage[utf8]{inputenc} %
\usepackage[T1]{fontenc}    %
\usepackage{hyperref}       %
\usepackage{url}            %
\usepackage{booktabs}       %
\usepackage{amsfonts}       %
\usepackage{nicefrac}       %
\usepackage{microtype}      %

\usepackage{amssymb,amsmath}%
\usepackage{longtable}
\usepackage{array}
\usepackage{xcolor}
\usepackage{algorithm}
\usepackage{algorithmic}
\usepackage{multirow}
\usepackage{subeqnarray}
\usepackage[numbers]{natbib} %
\usepackage{setspace} %
\usepackage{mathtools}
\usepackage{IEEEtrantools}
\usepackage{subcaption}
\usepackage{wrapfig,lipsum,booktabs}
\usepackage{pdfpages}

\renewcommand{\vec}{\boldsymbol}

\DeclarePairedDelimiterX{\expectarg}[1]{[}{]}{%
  \ifnum\currentgrouptype=16 \else\begingroup\fi
  \activatebar#1
  \ifnum\currentgrouptype=16 \else\endgroup\fi
}
\newcommand{\pprob}{\operatorname{p}\probarg}
\DeclarePairedDelimiterX{\probarg}[1]{(}{)}{%
  \ifnum\currentgrouptype=16 \else\begingroup\fi
  \activatebar#1
  \ifnum\currentgrouptype=16 \else\endgroup\fi
}

\newcommand{\qprob}{\operatorname{q}\probargq}
\DeclarePairedDelimiterX{\probargq}[1]{(}{)}{%
  \ifnum\currentgrouptype=16 \else\begingroup\fi
  \activatebar#1
  \ifnum\currentgrouptype=16 \else\endgroup\fi
}
\newcommand{\innermid}{\nonscript\;\delimsize\vert\nonscript\;}
\newcommand{\activatebar}{%
  \begingroup\lccode`\~=`\|
  \lowercase{\endgroup\let~}\innermid
  \mathcode`|=\string"8000
}
\newcommand{\matr}[1]{\mathbf{#1}}
\RequirePackage{xcolor}

\newcommand{\expt}[2]{\mathbb{E}_{#1}\left[#2\right]}
\newcommand{\kl}[2]{\text{KL}\left[#1\parallel#2\right]}

\newcommand{\defeq}{\vcentcolon=}

\newcommand{\bcdot}{\boldsymbol{\cdot}}

\newcommand{\cbdot}{\boldsymbol{\cdot}}

\usepackage{xr}
\title{Variational Autoencoders for Sparse and Overdispersed Discrete Data}

\author{He Zhao\IEEEauthorrefmark{1}\thanks{he.zhao@monash.edu} \And Piyush Rai\IEEEauthorrefmark{2}\\ \And Lan Du\IEEEauthorrefmark{1} \And Wray Buntine\IEEEauthorrefmark{1}  \And Mingyuan Zhou\IEEEauthorrefmark{3}\\\\
\IEEEauthorrefmark{1}Monash University, Australia
\\
\IEEEauthorrefmark{2}Indian Institute of Technology, Kanpur, India
\\
\IEEEauthorrefmark{3}The University of Texas at Austin, USA
}

\begin{document}

\maketitle

\begin{abstract}
Many applications, such as text modelling, high-throughput sequencing, and recommender systems, require analysing sparse, high-dimensional, and overdispersed discrete (count-valued or binary) data. 
Although probabilistic matrix factorisation and linear/nonlinear latent factor models have enjoyed great success in modelling such data, many existing models may have inferior modelling performance due to the insufficient capability of modelling overdispersion in count-valued data and model misspecification in general.
In this paper, we comprehensively study these issues and propose a variational autoencoder based framework that generates discrete data via negative-binomial distribution. We also examine the model's ability to capture properties, such as self- and cross-excitations in discrete data, which is critical for modelling overdispersion. We conduct extensive experiments on three important problems from discrete data analysis: text analysis, collaborative filtering, and multi-label learning. Compared with several state-of-the-art baselines, the proposed models achieve significantly better performance on the above problems.

\end{abstract}

\section{Introduction}
\label{sec-intro}

Discrete data are ubiquitous in many applications.
For example, in text analysis, a collection of documents can be represented 
as a word-document count matrix with the bag-of-words assumption;
in recommender systems,
users' shopping history can be represented as a binary (or count) item-user matrix, with each entry indicating whether or not a user has bought an item (or its purchase count);
in extreme multi-label learning problems, data samples can be tagged with a large set of labels, presented by a binary label matrix. Such kinds of data are often characterised by high-dimensionality and extreme sparsity. %

With the ability to handle high-dimensional and sparse matrices, Probabilistic Matrix Factorisation (PMF)~\citep{mnih2008probabilistic} has been a key method of choice for such problems.
PMF assumes data is generated from a suitable probability distribution, %
parameterised by low-dimensional latent factors.
When it comes to discrete data,
Latent Dirichlet Allocation (LDA)~\citep{blei2003latent} and Poisson Factor Analysis (PFA)~\citep{canny2004gap,zhou2012beta} are the two representative models that generate data samples using the multinomial and Poisson distributions, respectively. 
Originally, LDA and PFA can be seen as single-layer models, whose modelling expressiveness may be limited. Therefore, extensive research has been devoted to extending them with hierarchical Bayesian priors~\citep{blei2010nested,paisley2015nested,gan2015learning,zhou2016augmentable}.
However, increasing model complexity with hierarchical priors can also complicate inference, and lack of scalability hinders their usefulness in analysing large-scale data.
The recent success of deep generative models such as Variational Autoencoders (VAEs)~\citep{kingma2013auto,rezende2014stochastic} on modelling real-valued data such as images has motivated machine learning practitioners to adapt VAEs to 
dealing with discrete data as done in recent works~\citep{miao2016neural, miao2017discovering,krishnan2018challenges,liang2018variational}. 
Instead of using the Gaussian distribution as the data distribution for real-valued data,
the multinomial distribution has been used for discrete data~\citep{miao2016neural,krishnan2018challenges,liang2018variational}. 
Following~\citet{liang2018variational}, we refer to these VAE-based models as ``MultiVAE'' (Multi for multinomial)\footnote{In terms of the generative process (encoder), the models in~\citet{miao2016neural,krishnan2018challenges,liang2018variational} are similar, despite that the inference procedures are different.}.
MultiVAE can be viewed as a deep nonlinear PMF
model, where the nonlinearity is introduced by the deep neural networks in the decoder.
Compared with conventional hierarchical Bayesian models, MultiVAE increases its modelling capacity without sacrificing the scalability, because of the use of amortized variational inference (AVI)~\citep{rezende2014stochastic}.

Nevertheless, the use of the multinomial data distribution in existing VAE models such as MultiVAE can lead to inferior modelling performance on discrete data due to: \textbf{1)} insufficient capability of modelling overdispersion in count-valued data, and \textbf{2)} model misspecification in binary data. Specifically, \textit{overdispersion}  (i.e., variance larger than the mean) describes the phenomenon that the data variability is %
large, which is a key property for large-scale count-valued data. For example, overdispersion in text data can behave as \textit{word burstiness}~\citep{church1995poisson,madsen2005modeling,doyle2009accounting,buntine2014experiments}, %
which happens as
if a %
word is seen in a document, it
may excite both itself and related ones.
Burstiness can cause overdispersion in text data because a document usually has a few bursty words occurring multiple times while other words only show up once or never, resulting in high variance in bag-of-word matrices.
Shown in~\citet{zhou2018nonparametric}, the deep-seated causes of insufficient capability of modelling overdispersion in existing PMF models with Poisson or multinomial are due to their limited ability of handling \textit{self-} and \textit{cross-excitation}~\citep{zhou2018nonparametric}.
Specifically, in the text data example, self-excitation captures the effect that if a word occurs in a document, it is likely to occur more times in the same document. On the other hand, cross-excitation models the effect that if a word such as ``puppy'' occurs, it will likely to excite the occurrences of related words such as ``dog.''
It can be shown that existing PMF models with Poisson/multinomial do not distinguish between self- and cross-excitation and usually assume that data are independently generated.
Moreover, besides count-valued data, binary-valued observations are also prevalent in many applications, such as in collaborative filtering and graph analysis.
It may not be proper to directly apply multinomial or Poisson to binary data, which is a common misspecification in many existing models.
This is because multinomial and Poisson may assign more than one count to one position, ignoring the fact that the data are binary. The misspecification could result in inferior modelling performance~\citep{zhou2015infinite}.

In this paper, we show the above two issues on modelling discrete data can be addressed in a principled manner using the negative-binomial (NB) distribution as the data distribution in a deep-structured PMF model. Specifically, in Section~\ref{sec-model}, we analytically demonstrate that using NB as the data likelihood in PMFs can explicitly capture self-excitation and using deep structures can provide more model capacity for capturing cross-excitation. Therefore, a deep PMF with NB is able to better handle overdispersion by sufficiently capturing both %
kinds of excitations. On the other hand, the usage of NB instead of multinomial enables us to develop a link function between the Bernoulli and negative-binomial distributions, which gives us the ability to handle binary data with superior modelling performance.
Beside the in-depth analytical study, we propose a deep PMF model called \textbf{N}egative-\textbf{B}inomial \textbf{V}ariational \textbf{A}uto\textbf{E}ncoder (\textbf{NBVAE} for short), a VAE-based framework generating data with a negative-binomial distribution. Extensive experiments have been conducted on three important problems of discrete data analysis: text analysis on bag-of-words data, collaborative filtering on binary data, and multi-label learning. Compared with several state-of-the-art baselines, NBVAE achieves significantly better performance on the above problems.

\section{Analytical study and model details}
\label{sec-model}
In this section,
we start with the introduction of our proposed NBVAE model for count-valued data, and then give a detailed analysis on how self- and cross-excitations are captured in different models and why NBVAE is capable of better handling them.
Finally, we describe the variants of NBVAE for modelling binary data and for multi-label learning, respectively.

\subsection{Negative-binomial variational autoencoder (NBVAE)}
\label{sec-nbvae}
Like the standard VAE model, NBVAE consists of two major components: the decoder for the generative process and the encoder for the inference process. Here we focus on the generative process and discuss the inference procedure in Section~\ref{sec-vi}.
Without loss of generality, we present our model in the case of bag-of-word data for a text corpus, but the model can generally work with any kind of count-valued matrices.
Suppose the bag-of-word data are stored in a $V$ by $N$ count matrix $\matr{Y} \in \mathbb{N}^{V \times N} =[\vec{y}_1,\cdots,\vec{y}_N]$, where $\mathbb{N} = \{0,1,2,\cdots\}$ and $V$ and $N$ are the number of documents and the size of the vocabulary, respectively.
To generate the occurrences of the words for the $j^{\text{th}}$ ($j \in \{1,\cdots N\}$) document, $\vec{y}_j \in \mathbb{N}^{V}$, we draw a $K$ dimensional latent representation $\vec{z}_j \in \mathbb{R}^{K}$ from a standard multivariate normal prior.
After that, $\vec{y}_j$ is drawn from a (multivariate) negative-binomial distribution with $\vec{r}_j \in \mathbb{R}_{+}^{V}$ ($\mathbb{R}_{+} = \{x: x \ge 0\}$) and $\vec{p}_j \in (0,1)^{V}$ as the parameters.
Moreover, $\vec{r}_j$ and $\vec{p}_j$ are obtained by transforming $\vec{z}_j$ from two nonlinear functions, $f_{\theta^{r}}(\bcdot)$ and $f_{\theta^{p}}(\bcdot)$, parameterised by $\theta^r$ and $\theta^p$, respectively.
The above generative process of $\pprob{\vec{y}_j | \vec{z}_j}$ can be formulated as follows:
\begin{IEEEeqnarray}{+rCl+x*}
\vec{z}_j \sim \mathcal{N}(\vec{0},\textbf{I}_{{K}}),
\vec{r}_j = \text{exp}\left(f_{\theta^r}(\vec{z}_j)\right),
\vec{p}_j = 1/(1 + \text{exp}\left(-f_{\theta^p}(\vec{z}_j\right)),
\vec{y}_{j} \sim \text{NB}(\vec{r}_j, \vec{p}_j).
\end{IEEEeqnarray}

In the above model, the output of $f_{\theta^p}(\bcdot)$, $\vec{p}_j$, is a $V$ dimensional vector.
Alternatively, if we set the output of $f_{\theta^p}(\bcdot)$ a single number $p_j \in (0,1)$ specific to document $j$,
according to~\citet[Theorem 1]{zhou2018nonparametric}, we can derive an alternative representation of NBVAE:
\begin{IEEEeqnarray}{+rCl+x*}
&\vec{z}_j \sim \mathcal{N}(\vec{0},\textbf{I}_{{K}}),
\vec{r}_j = \text{exp}\left(f_{\theta^r}(\vec{z}_j)\right),
p_j = 1/(1 + \text{exp}\left(-f_{\theta^p}(\vec{z}_j\right)),&\\
&y_{\cbdot j} \sim \text{NB}\left(r_{\cbdot j}, p_j\right),
\label{eq-dmvae}
\vec{y}_j \sim \text{DirMulti}\left(y_{\cbdot j}, \vec{r}_{j}\right),&
\end{IEEEeqnarray} where ``DirMulti'' stands for the Dirichlet-multinomial distribution, $y_{\cbdot j} = \sum_{v}^{V} y_{vj}$ is the total number of words of document $j$, and $r_{\cbdot j} = \sum_{v}^{V} r_{vj}$.
Accordingly, we refer this representation of NBVAE to as NBVAE$_{\text{dm}}$ (dm for Dirichlet-multinomial).
Note that NBVAE$_{\text{dm}}$ can also be viewed as a deep nonlinear generalization of models based on Dirichlet-multinomial to capture word burstiness~\citep{doyle2009accounting,buntine2014experiments}. 
Compared with NBVAE, when doing inference for NBVAE$_{\text{dm}}$, we can treat $y_{\bcdot j}$, i.e., the total number of words as an observed variable.

\subsection{How NBVAE captures self- and cross-excitations}
\label{sec-self-cross}
We now compare NBVAE and other PMF models in terms of their ability in capturing self- and cross-excitations in count-valued data. 
For easy comparison, we present the related PMF models with a unified framework.
Without loss of generality, we demonstrate the framework in the case of bag-of-word text data, where the $i^{\text{th}}$ word's type in document $j$ is $w^{i}_{j} \in \{1,\cdots,V\}$ and the occurrence of $v$ in document $j$ is $y_{vj}$.
The first thing we are interested in is the data distribution that generates the word occurrences of document $j$.
Specifically, we can generate $\{w^{i}_{j}\}_{i=1}^{y_{\bcdot j}}$ from $\pprob{w^{i}_j = v| \vec{l}_j} \propto l_{vj} / l_{\bcdot j}$ where $\vec{l}_j \in \mathbb{R}_{+}^V$ is the model parameter. After all the word types are generated, we can count the occurrences of different types of words by $y_{vj} = \sum_{i}^{y_{\bcdot j}} \textbf{1}(w^i_j = v)$, where $\textbf{1}(\bcdot)$ is the indicator function.
Alternatively, we can directly generate $\vec{y}_{j}$ from $\pprob{\vec{y}_j | \vec{l}_j}$. 
As shown later, for PMF models, $\vec{l}_j$ explicitly or implicitly takes a factorised form.
Moreover, we are more interested in the predictive distribution of a word, $w^{i}_j$, conditioned on the other words' occurrences in the corpus, $\matr{Y}^{-ij}$, which can be presented as follows:
\begin{IEEEeqnarray}{+rCl+x*}
\pprob{w^{i}_j = v | \matr{Y}^{-ij}} &\propto& \int \pprob{w^{i}_j = v | \vec{l}'_j} \pprob{\vec{l}'_j | \matr{Y}^{-ij}} \text{d} \vec{l}'_j = \expt{\pprob{\vec{l}'_j | \matr{Y}^{-ij}}}{\pprob{w^{i}_j = v | \vec{l}'_j}},
\end{IEEEeqnarray} where $\vec{l}'_j$ is the predictive rate computed with the parameters obtained from the posterior.

Now we reformulate the models related to NBVAE into the above framework, including Poisson Factor Analysis (PFA)~\citep{canny2004gap,zhou2012beta}, Latent Dirichlet Allocation (LDA)~\citep{blei2003latent}, MultiVAE~\citep{miao2016neural,krishnan2018challenges,liang2018variational}, and Negative-Binomial Factor Analysis (NBFA)~\citep{zhou2018nonparametric}, as follows:

\textbf{PFA: } It is obvious that PFA directly fits into this framework, where $\pprob{\vec{y}_j | \vec{l}_j}$ is the Poisson distribution and $\vec{l}_j = \matr{\Phi}\vec{\theta}_j$. 
Here $\matr{\Phi} \in \mathbb{R}_{+}^{V \times K} = [\vec{\phi}_1,\cdots,\vec{\phi}_K]$ is the factor loading matrix and $\matr{\Theta} \in \mathbb{R}_{+}^{K \times N}  = [\vec{\theta}_1,\cdots,\vec{\theta}_N]$ is the factor score matrix. 
Their linear combinations determine the probability of the occurrence of $v$ in document $j$.

\textbf{LDA: } Originally, LDA explicitly assigns a topic $z^{i}_{j} \in \{1,\cdots,K\}$ to $w^{i}_{j}$, with the following process: $z^{i}_{j} \sim \text{Cat}(\vec{\theta}_j / \theta_{\bcdot j})$ and $w^{i}_{j} \sim \text{Cat}(\vec{\phi}_{z_{ij}})$, where $\theta_{\bcdot j} = \sum_{k}^{K} \theta_{kj}$ and ``Cat'' is the categorical distribution.
By collapsing all the topics, we can derive an equivalent representation of LDA, in line with the general framework: $\vec{y}_j \sim \text{Multi}(y_{\bcdot j}, \vec{l}_j)$, where $\vec{l}_j = \matr{\Phi} \vec{\theta}_j / \theta_{\bcdot j}$.

\textbf{MultiVAE: } MultiVAE generates data from a multinomial distribution, whose parameters are constructed by the decoder:
 $\vec{y}_j \sim \text{Multi}(y_{\bcdot j}, \vec{l}_j)$, where $\vec{l}_j = \text{softmax}(f_{\theta}(\vec{z}_j))$. As shown in~\citet{krishnan2018challenges}, MultiVAE can be viewed as a nonlinear PMF model.

\textbf{NBFA: } NBFA uses a negative-binomial distribution as the data distribution, the generative process of which can be represented as: $\vec{y}_j \sim \text{NB}(\vec{l}_j, p_j)$, where $\vec{l}_j = \matr{\Phi} \vec{\theta}_j$.

\begin{table}[]
\caption{\small Comparison of the data distributions, model parameters, predictive rates, and posteriors. $\qprob{\bcdot}$ denotes the encoder in VAE models, which will be introduced in Section~\ref{sec-vi}.}
\label{tb-model}
\centering
\resizebox{0.999\linewidth}{!}{
\begin{tabular}{|c|l|l|l|l|}
\hline
\multicolumn{1}{|c|}{Model}     &  \multicolumn{1}{c|}{Data distribution} & \multicolumn{1}{c|}{Model parameter} & \multicolumn{1}{c|}{Predictive rate} & \multicolumn{1}{c|}{Posterior}\\ \hline\hline
PFA      &   $\vec{y}_j \sim \text{Poisson}(\vec{l}_j)$ &    $\vec{l}_j = \matr{\Phi}\vec{\theta}_j$     &     $l'_{vj} \propto \sum_{k}^{K} \phi_{vk} \theta_{kj}$   &   $\matr{\Phi}, \vec{\theta}_j \sim \pprob{\matr{\Phi}, \vec{\theta}_j | \matr{Y}^{-ij}}$   \\ \hline 
LDA      &   $\vec{y}_j \sim \text{Multi}(y_{\bcdot j}, \vec{l}_j)$                &   $\vec{l}_j = \matr{\Phi} \vec{\theta}_j / \theta_{\bcdot j}$    &     $l'_{vj} \propto \sum_{k}^{K} \phi_{vk} \theta_{kj} / \theta_{\bcdot j}$   &   $\matr{\Phi}, \vec{\theta}_j \sim \pprob{\matr{\Phi}, \vec{\theta}_j | \matr{Y}^{-ij}}$     \\ \hline
MultiVAE &   $\vec{y}_j \sim \text{Multi}(y_{\bcdot j}, \vec{l}_j)$                &   $\vec{l}_j = \text{softmax}(f_{\theta}(\vec{z}_j))$     &       $l'_{vj} \propto \text{softmax}(f_{\theta}(\vec{z}_j))_v$ &       $\vec{z}_j \sim \qprob{ \vec{z}_{j} | \matr{Y}^{-ij}}$           \\ \hline
NBFA     &   $\vec{y}_j \sim \text{NB}(\vec{l}_j, p_j)$                &        $\vec{l}_j = \matr{\Phi} \vec{\theta}_j$       &        $l'_{vj} \propto (y_{vj}^{-i} + \sum_{k}^{K} \phi_{vk} \theta_{kj}) p_j$     &    $\matr{\Phi}, \vec{\theta}_j, p_j \sim \pprob{\matr{\Phi}, \vec{\theta}_j, p_j | \matr{Y}^{-ij}}$  \\ \hline
NBVAE    &   $\vec{y}_j \sim \text{NB}(\vec{r}_j, \vec{p}_j)$                 & \begin{tabular}[c]{@{}l@{}}$\vec{r}_{j} = \text{exp}\left(f_{\theta^r}(\vec{z}_j)\right)$ \\ $\vec{p}_{j} = 1/(1 + \text{exp}\left(-f_{\theta^p}(\vec{z}_j\right))$\end{tabular}      &     $l'_{vj} \propto \frac{y_{vj}^{-i} + \text{exp}\left(f_{\theta^r}(\vec{z}_j)\right)_v}{\left(1 + \text{exp}\left(-f_{\theta^p}(\vec{z}_j\right)\right)_v}$      &       $\vec{z}_j \sim \qprob{ \vec{z}_{j} | \matr{Y}^{-ij}}$    \\ \hline
NBVAE$_{\text{dm}}$    &     $\vec{y}_j \sim \text{DirMulti}\left(y_{\cbdot j}, \vec{r}_{j}\right)$              &      \begin{tabular}[c]{@{}l@{}}$\vec{r}_{j} = \text{exp}\left(f_{\theta^r}(\vec{z}_j)\right)$ \\ $p_{j} = 1/(1 + \text{exp}\left(-f_{\theta^p}(\vec{z}_j\right))$\end{tabular}    &   $l'_{vj} \propto y_{vj}^{-i} + \text{exp}\left(f_{\theta^r}(\vec{z}_j)\right)_v$     &         $\vec{z}_j \sim \qprob{ \vec{z}_{j} | \matr{Y}^{-ij}}$   \\ \hline
\end{tabular}
}
\end{table}

\begin{wraptable}{r}{0.6\linewidth}
\caption{\small How different models capture self- and cross-excitation}
\label{tb-sc}
\resizebox{0.55\textwidth}{!}{
\begin{tabular}{|c|c|c|}
\hline
Model       & Self-excitation   & Cross-excitation             \\ \hline\hline
PFA         & \multicolumn{2}{c|}{Single-layer structure}      \\ \hline
LDA         & \multicolumn{2}{c|}{Single-layer structure}      \\\hline
MultiVAE    & \multicolumn{2}{c|}{Multi-layer neural networks} \\ \hline
NBFA        &     $y_{vj}^{-i}$              & Single-layer structure       \\ \hline

\begin{tabular}[c]{@{}l@{}}\textbf{NBVAE}\\ \textbf{NBVAE$_{\text{dm}}$}\end{tabular}
&   $y_{vj}^{-i}$                & Multi-layer neural networks  \\ \hline
\end{tabular}
}
\end{wraptable}

The above comparisons on the data distributions and predictive distributions of those models are shown in Table~\ref{tb-model}.
In particular, we can show a model's capacity of capturing self- and cross-excitations by analysing its predictive distribution.
Note that $y_{vj}^{-i}$ denotes the number of $v$'s occurrences in document $j$ excluding the $i^{\text{th}}$ word.
If we compare PFA, LDA, MultiVAE V.S. NBFA, NBVAE, NBVAE$_{\text{dm}}$, it can be seen that the latter three models with NB as their data distributions explicitly capture self-excitation via the term $y_{vj}^{-i}$ in the predictive distributions. Specifically, if $v$ appears more in document $j$, $y_{vj}^{-i}$ will be larger, leading to larger probability that $v$ shows up again. That is to say, the latter three models capture word burstiness directly with $y_{vj}^{-i}$.
However, PFA, LDA, and MultiVAE cannot capture self-excitation directly because they predict a word purely based on the interactions of the latent representations and pay less attention to the existing word frequencies. %
Therefore, even with deep structures, their potential of modelling self-excitation is still limited.
Moreover, for the models with NB, i.e., NBVAE and NBFA, as self-excitation is explicitly captured by $y_{vj}^{-i}$,
the interactions of the latent factors are only responsible for cross-excitation. 
Specifically, NBFA applies a single-layer linear combination of the latent representations, i.e., $\sum_{k}^{K} \phi_{vk} \theta_{kj}$, while NBVAE can be viewed as a deep extension of NBFA, using a deep neural network to conduct multi-layer nonlinear combinations of the latent representations, i.e, $r_{vj} = \text{exp}\left(f_{\theta^r}(\vec{z}_j)\right)_v$ and $p_{vj} = 1/(1 + \text{exp}\left(-f_{\theta^p}(\vec{z}_j\right))_v$. Therefore, NBVAE enjoys richer modelling capacity than NBFA on capturing cross-excitation.
Finally, we summarise our analysis on how self- and cross-excitations are captured in related models in Table~\ref{tb-sc}.

\subsection{NBVAE for binary data} 

In many problems, discrete data are binary-valued. 
For example, suppose the binary matrix $\matr{Y} \in \{0,1\}^{V \times N}$ stores the buying history of $N$ users on $V$ items, where $y_{vj}=1$ indicates that user $j$ has brought item $v$.
Precious models like MultiVAE~\citep{krishnan2018challenges,liang2018variational} treat such binary data as counts, which is a model misspecification that is likely to result in inferior performance.
To better model binary data, we develop a simple yet effective method that links NBVAE and the Bernoulli distribution.
Specifically, inspired by the link function used in~\citet{zhou2015infinite}, we first generate a latent discrete intensity vector, $\vec{m}_j \in \mathbb{N}^{V}$, from the generative process of NBVAE, where $m_{vj}$ can be viewed as the interest of user $j$ on item $v$. Next, we generate the binary buying history of user $j$ by thresholding the discrete intensity vector at one, as follows:
\begin{IEEEeqnarray}{+rCl+x*}
\vec{m}_j \sim \text{NBVAE}(\vec{z}_j),~~
\vec{y}_j = \textbf{1}(\vec{m}_j \ge 1).
\end{IEEEeqnarray}
As $\vec{m}_j$ is drawn from NB, we do not have to explicitly generate it. Instead, if we marginalise it out, we can get the following data likelihood: $\vec{y}_j \sim \text{Bernoulli}\left(1 - (1-\vec{p}_{j})^{\vec{r}_{j}}\right),$
where $\vec{r}_j$ and $\vec{p}_j$ have the same construction of the original NBVAE. Here we refer to this extension of NBVAE as NBVAE$_{\text{b}}$ (b for binary).
Given the fact that the NB distribution is a gamma mixed Poisson distribution, %
the elements of $\vec{m}_j$ can be viewed to be individually generated from a Poisson distribution. In contrast, a vector's elements are jointly generated from multinomial in MutiVAE and Dirichlet-multinomial in NBVAE$_{\text{dm}}$. Therefore, the link function is inapplicable to them.

\subsection{NBVAE for multi-label learning}

Inspired by the appealing capacity of NBVAE$_{\text{b}}$ for modelling binary data and the idea of Conditional VAE~\citep{NIPS2015_5775}, we develop a conditional version of NBVAE, named NBVAE$_{\text{c}}$ (c for conditional), for extreme multi-label learning.
Being increasingly important recently, multi-label learning is a supervised task where there is an extremely large set of labels while an individual sample is associated with a small subset of the labels.
Specifically, suppose there are $N$ samples, each of which is associated with a $D$ dimensional feature vector $\vec{x}_j \in \mathbb{R}^{D}$ and a binary label vector $\vec{y}_j \in \{0,1\}^V$. $V$ is the number of labels that can be very large, and $y_{vj}=1$ indicates sample $j$ is labelled with $v$. The goal is to predict the labels of a sample given its features. Here the label matrix $\matr{Y} \in \{0,1\}^{V \times N}$ is a large-scale, sparse, binary matrix. 
The general idea is that instead of drawing the latent representation of a sample from an uninformative prior (i.e., standard normal) in NBVAE$_{\text{b}}$, we use an informative prior constructed with the sample's feature in NBVAE$_{\text{c}}$.
Specifically, 
we introduce a parametric function $f_{\psi}(\cdot)$ that transforms the features of sample $j$ to the mean and variance of the normal prior, formulated as follows:
\begin{IEEEeqnarray}{+rCl+x*}
\label{eq-z-prior-1}
\vec{\mu}_j, \vec{\sigma}^2_j \defeq f_{\psi}(\vec{x}_j),~~
\label{eq-z-prior-2}
\vec{z}_j \sim \mathcal{N}(\vec{\mu}_j,\text{diag}\{\vec{\sigma}^2_j\}),~~
\vec{y}_{j} \sim \text{NBVAE$_{\text{b}}$}(\vec{z}_j).
\end{IEEEeqnarray}
Note that $f_{\psi}(\cdot)$ defines $\pprob{\vec{z}_j | \vec{x}_j}$, which
encodes the features of a sample into the prior of its latent representation. Therefore, it is intuitive to name it the \textit{feature encoder}. With the above construction, given the feature vector of a testing sample $j^*$, we can feed $\vec{x}_{j^*}$ into the feature encoder to sample the latent representation, $\vec{z}_{j^*}$, then feed it into the decoder to predict its labels.

\section{Variational inference}
\label{sec-vi}
The inference of NBVAE, NBVAE$_{\text{dm}}$, and NBVAE$_{\text{b}}$ follows the standard amortized variational inference procedure of VAEs, where instead of directly deriving the posterior of a model $\pprob{\vec{z}_j | \vec{y}_j}$, we propose a data-dependent variational distribution $\qprob{\vec{z}_j | \vec{y}_j}$ (i.e., encoder) to approximate the true posterior,
constructed as follows:
\begin{IEEEeqnarray}{+rCl+x*}
\label{eq-z-posterior-1}
\vec{\widetilde{\mu}}_j, \vec{\widetilde{\sigma}}^2_j = f_{\phi}(\vec{y}_j),
\label{eq-z-posterior-2}
\vec{z}_j \sim \mathcal{N}(\vec{\widetilde{\mu}}_j,\text{diag}\{\vec{\widetilde{\sigma}}^2_j\}).
\end{IEEEeqnarray}
Given $\qprob{\vec{z}_j | \vec{y}_j}$, the learning objective is to maximise the Evidence Lower BOund (ELBO) of the marginal likelihood of the data, i.e., $\expt{\qprob{\vec{z}_j | \vec{y}_j}}{\log{\pprob{\vec{y}_j | \vec{z}_j}}} - \kl{\qprob{\vec{z}_j | \vec{y}_j}}{\pprob{\vec{z}_j}}$, in terms of the decoder parameters $\theta^r$, $\theta^p$ and the encoder parameter $\phi$.
Here the reparametrization trick~\citep{kingma2013auto,rezende2014stochastic} is used to sample from $\qprob{\vec{z}_j | \vec{y}_j}$.

The difference in the inference NBVAE$_{\text{c}}$ and the above models is: because of the use of the informative prior constructed with sample features, the Kullback-Leiber (KL) divergence on the RHS of the ELBO is calculated between two non-standard multivariate normal distributions, i.e. $\kl{\qprob{\vec{z}_j | \vec{y}_j}}{\pprob{\vec{z}_j | \vec{x}_j}}$. Moreover, the feature encoder, $f_{\psi}(\cdot)$, is also learned as it is involved in $\pprob{\vec{z}_j | \vec{x}_j}$. 
Another heuristic modification to the inference of NBVAE$_{\text{c}}$ is that instead of always drawing $\vec{z}_j$ from the encoder, i.e., $\vec{z}_j \sim \qprob{\vec{z}_j | \vec{y}_j}$, we draw $\vec{z}_j$ from the feature encoder, i.e., $\vec{z}_j \sim \pprob{\vec{z}_j | \vec{x}_j}$ in every other iteration. The modification is interesting and intuitive. Given the ELBO of NBVAE$_{\text{c}}$, the feature encoder only contributes to the KL divergence serving as the regularisation term and is not directly involved in the generation of data (labels). Recall that without knowing any labels of a test sample, we cannot use the encoder and the feature encoder is the key part to get the latent representation of the sample. Therefore, the above modification enables the feature encoder to directly contribute to the generation of labels, which improves its performance in the testing phase.
We give more analysis and empirically demonstrations in the appendix.

\section{Related work}

\textbf{Probabilistic matrix factorisation models for discrete data.}
Lots of well-known  models fall into this category, including LDA~\citep{blei2003latent} and PFA~\citep{zhou2012beta}, as well as their hierarchical extensions such as Hierarchical Dirichlet Process~\citep{teh2012hierarchical}, nested Chinese Restaurant Process (nCRP)~\citep{blei2010nested}, nested Hierarchical Dirichlet Process (nHDP)~\citep{paisley2015nested}, Deep Poisson Factor Analysis (DPFA)~\citep{gan2015scalable}, Deep Exponential Families (DEF)~\citep{ranganath2015deep}, 
Deep Poisson Factor Modelling (DPFM)~\citep{henao2015deep}, and Gamma Belief Networks (GBNs)~\citep{zhou2016augmentable}.
Among various models, the closest ones to ours are NBFA~\citep{zhou2018nonparametric} and non-parametric LDA (NP-LDA)~\citep{buntine2014experiments}, which generate data with the negative-binomial distribution and Dirichlet-multinomial distribution, respectively.
Our models can be viewed as a deep generative extensions to them, providing better model expressiveness, flexibility, as well as inference scalability.

\textbf{VAEs for discrete data.} 
\citet{miao2016neural} proposed the Neural Variational Document Model (NVDM), which extended the standard VAE with multinomial likelihood for document modelling and \citet{miao2017discovering} further built a VAE to generate the document-topic distributions in the LDA framework.
\citet{srivastava2016autoencoding} developed an AVI algorithm for the inference of LDA, which can be viewed as a VAE model.
\citet{card2018neural} introduced a general VAE framework for topic modelling with meta-data.
\citet{gronbech2019scvae} recently proposed a Gaussian mixture VAE with negative-binomial for gene data, which has a different construction to ours and the paper does not consider binary data, multi-label learning, or in-depth analysis.
\citet{krishnan2018challenges} recently found that using the standard training algorithm of VAEs in large sparse discrete data may suffer from model underfitting and proposed a stochastic variational inference (SVI)~\citep{hoffman2013stochastic} algorithm initialised by AVI to mitigate this issue.
In the collaborative filtering domain, \citet{liang2018variational} noticed a similar issue and alleviated it by proposing MultiVAE with a training scheme based on KL annealing~\citep{bowman2016generating}.
Note that NVDM, NFA, and MultiVAE are the closest ones to ours, their generative processes are very similar but their inference procedures are different. NFA is reported to outperform NVDM on text analysis~\citep{krishnan2018challenges} while MultiVAE is reported to have better performance than NFA on collaborative filtering tasks~\citep{liang2018variational}.
Compared with them, we improve the modelling performance in a different way, i.e., by better capturing self- and cross-excitations so as to better handle overdispersion.  %
Moreover, NFA and MultiVAE use the multinomial distribution, which may not work properly for binary data.
To our knowledge, the adaptation of VAEs to the multi-label learning area is rare, because modelling large-scale sparse binary label matrices may hinder the use of most existing models. In terms of the way of incorporating features, our NBVAE$_\text{c}$ is related to Conditional VAE~\cite{NIPS2015_5775}. However, several adaptations have been made in our model to get the state-of-the-art performance in multi-label learning.

\section{Experiments}
In this section, we evaluate the proposed models on three important applications of discrete data: text analysis, collaborative filtering, and multi-label learning with large-scale real-world datasets. In the experiments, we ran our models multiple times and report the average results. The details of the datasets, experimental settings, evaluation metrics, and more in-depth experiments are shown in the appendix.

\subsection{Experiments on text analysis}
Our first set of experiments is on text analysis.
We used three widely-used corpora~\citep{srivastava2013modeling,gan2015scalable,henao2015deep,cong2017deep}:
20 News Group (20NG), Reuters Corpus Volume (RCV), and Wikipedia (Wiki).
The details of these datasets are shown in the appendix.
For the evaluation metric, following~\citet{wallach2009}
we report per-heldout-word perplexity of all the models, 
which is a widely-used metric for text analysis. Note that we used the same perplexity calculation for all the compared models, detailed in the appendix.
We compared our proposed NBVAE and NBVAE$_{\text{dm}}$ with the following three categories of state-of-the-art models for text analysis: 
\textbf{1)} Bayesian deep extensions of PFA and LDA: DLDA~\citep{cong2017deep}, DPFM~\citep{henao2015deep}, DPFA~\citep{gan2015learning} with different kinds of inference algorithms such as Gibbs sampling, stochastic variational inference, and stochastic gradient MCMC (SGMCMC)~\citep{chen2014stochastic};
\textbf{2)} NBFA~\citep{zhou2018nonparametric}, is a recently-proposed single-layer factor analysis model with negative-binomial likelihood, whose inference is done by Gibbs sampling. 
As NBFA is a nonparameric model, we used its truncated version to compare with other models. 
\textbf{3)} MultiVAE~\citep{liang2018variational,krishnan2018challenges}, a recent VAE model for 
discrete data with the multinomial distribution as the data distribution. 
We used the original implementation of MultiVAE~\citep{liang2018variational}. The details of the experimental settings are shown in Section~\ref{apx-sec-setup-text}.

\begin{table}[]
\centering
\caption{\small Perplexity comparisons. ``Layers'' indicate the architecture of the hidden layers (for VAE models, it is the hidden layer architecture of the encoder.). Best results for each dataset are in boldface. TLASGR and SGNHT are the algorithms of SGMCMC, detailed in the papers of DLDA~\citep{cong2017deep} and DPFA~\citep{gan2015scalable}. Some results of the models with Gibbs sampling on RCV and Wiki are not reported because of the scalability issue.}
\label{tb-pp}
\resizebox{0.55\linewidth}{!}{
\begin{tabular}{|c|c|l|c|c|c|}
\hline
Model          & Inference & \multicolumn{1}{c|}{Layers}      & 20NG & RCV & Wiki \\ \hline \hline
DLDA           & TLASGR    & 128-64-32 & 757  & 815 & 786   \\ 
DLDA           & Gibbs     & 128-64-32 & 752  & 802 & - \\ 
DPFM           & SVI       & 128-64    & 818  & 961 & 791\\ 
DPFM           & MCMC      & 128-64    & 780  & 908 & 783  \\ 
DPFA-SBN       & Gibbs     & 128-64-32 & 827  & - & - \\ 
DPFA-SBN       & SGNHT     & 128-64-32 & 846  & 1143 & 876 \\ 
DPFA-RBM       & SGNHT     & 128-64-32 & 896  & 920 & 942\\ \hline \hline
NBFA & Gibbs     & 128       & 690  & 702  & -  \\ \hline \hline
MultiVAE       & VAE       & 128-64    & 746  & 632  &  629 \\ 
MultiVAE       & VAE       & 128       & 772  & 786  &  756\\ \hline \hline
NBVAE$_{\text{dm}}$    & VAE       & 128-64    & \textbf{678}  & 590  & 475\\ 
NBVAE$_{\text{dm}}$    & VAE       & 128       & 749  & 709 & 526\\ 
NBVAE          & VAE       & 128-64    & 688  & \textbf{579}  & \textbf{464}\\
NBVAE          & VAE       & 128       & 714  & 694 & 529\\ \hline
\end{tabular}
}
\end{table}

The perplexity results are shown in Table~\ref{tb-pp}. Following~\citet{gan2015scalable,henao2015deep,cong2017deep}, we report the performance of DLDA, DPFM, and DPFA with two and/or three hidden layers, which are the best results reported in their papers.
For the VAE-based models, we varied the network architecture with one and two hidden layers 
and varied the depths and widths of the layers, as shown in Table~\ref{tb-pp}.
We have the following remarks on the results:
\textbf{1)} If we compare NBFA with the deep Bayesian models with Poisson distributions listed above it in the table, 
the results show that modelling self-excitation with the negative-binomial distribution in NBFA has a large contribution to the modelling performance.
\textbf{2)} It can be observed that the single-layer VAE models (i.e., MultiVAEs with one layer) achieve no better results than NBFA. 
However, when multi-layer structures were used, VAE models largely improve their performance.
This shows the increased model capacity with deeper neural networks is critical to getting better modelling performance via cross-excitation.
\textbf{3)} Most importantly, our proposed NBVAE and NBVAE$_\text{dm}$ significantly outperform 
all the other models, which proves the necessity of modelling self-excitation explicitly and modelling cross-excitation with the deep structures of VAE.
\textbf{4)} Furthermore, the differences of perplexity between NBVAE and NBVAE$_\text{dm}$ are quite marginal, which is in line with the fact that they are virtually equivalent representations of the same model. 
To further study why our models achieve better perplexity results than the others, 
we conducted an additional set of comprehensive experiments with in-depth analysis in Section~\ref{apx-sec-text-more-exp} of the appendix.

\begin{figure*}[t]
        \centering
         \begin{subfigure}[b]{0.24\linewidth}
                 \centering
                 \caption{ML-10M}
                 \includegraphics[width=0.99\textwidth]{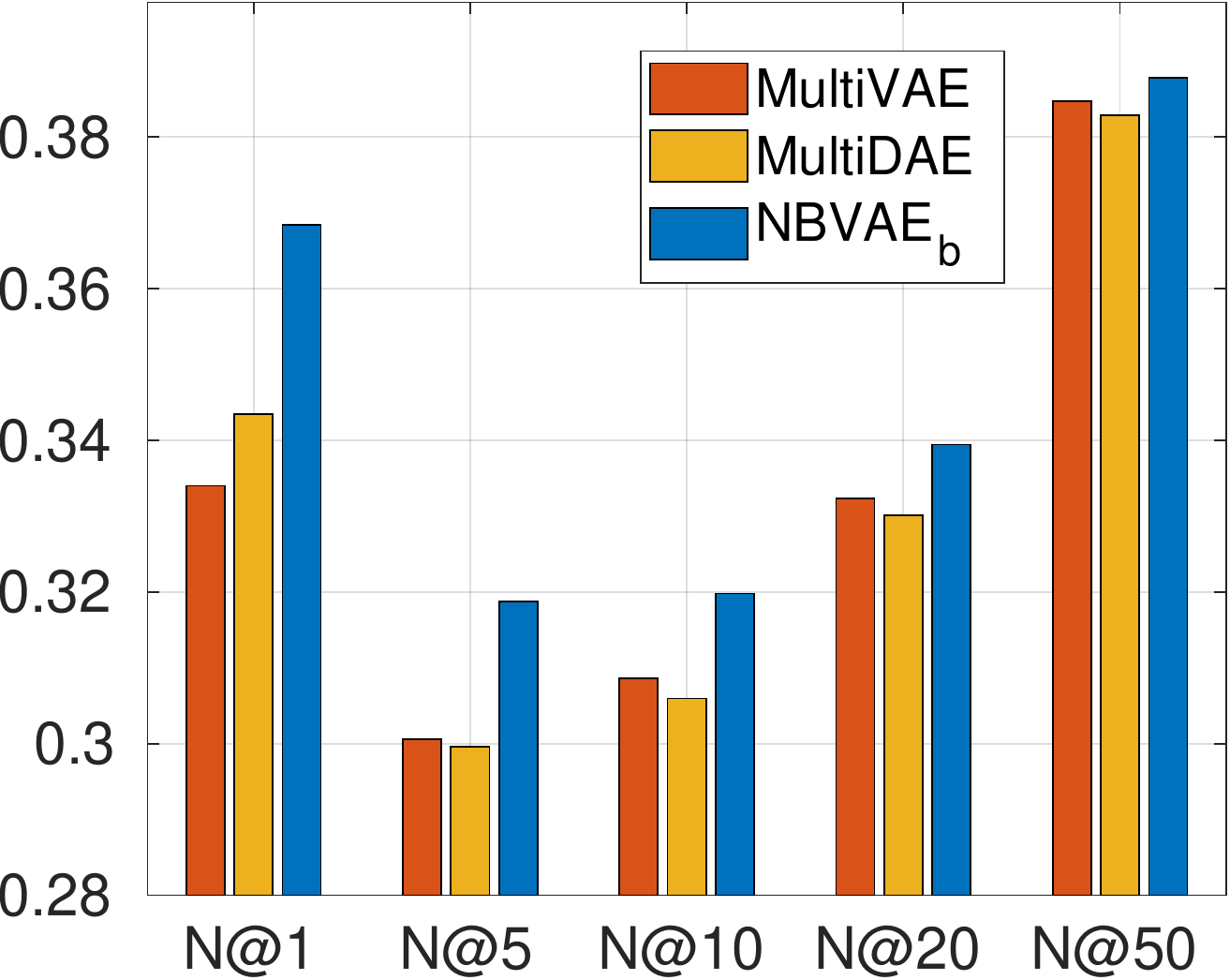}
         \end{subfigure}
         \begin{subfigure}[b]{0.24\linewidth}
                 \centering
                 \caption{ML-20M}
                 \includegraphics[width=0.99\textwidth]{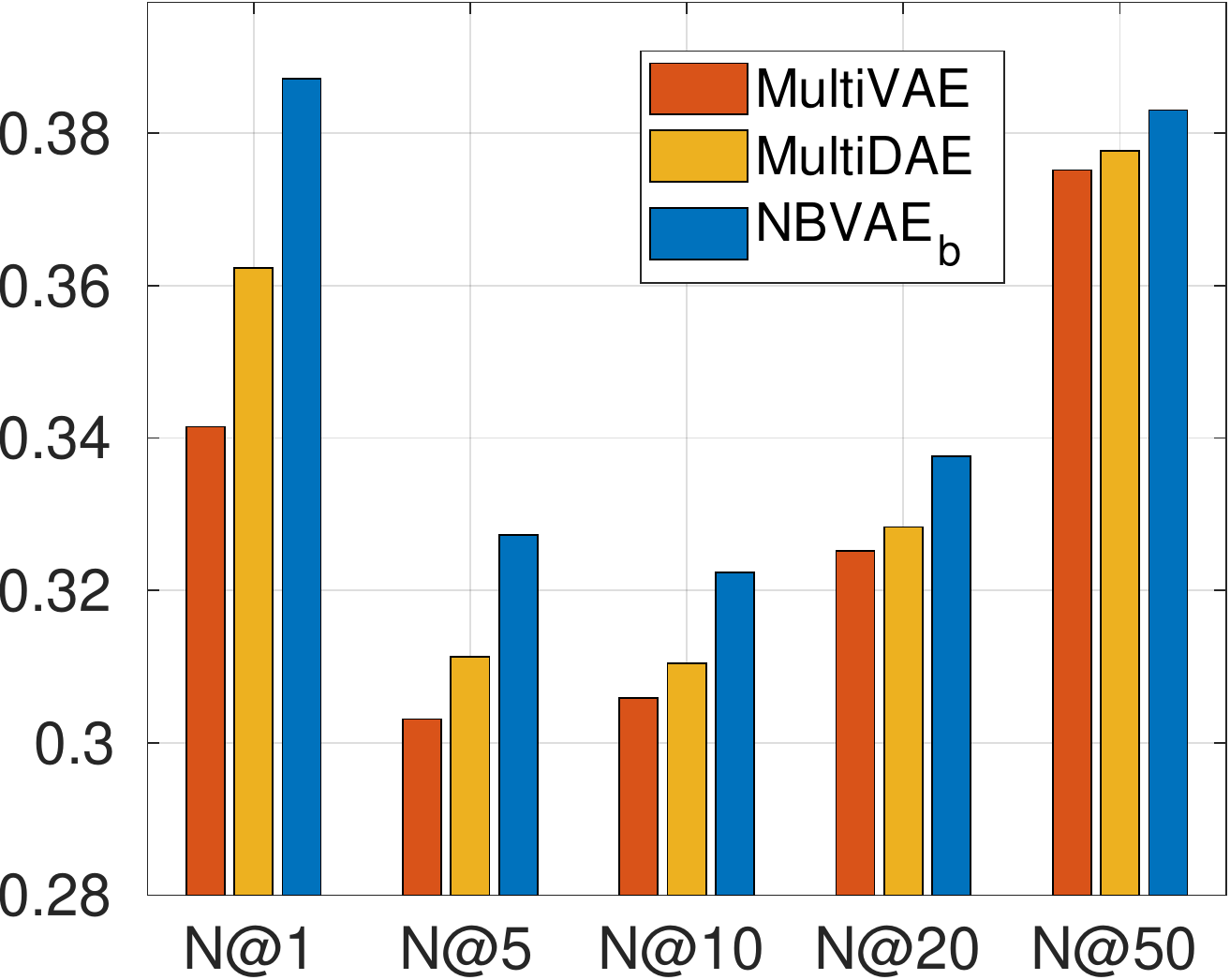}
         \end{subfigure}%
         \begin{subfigure}[b]{0.24\linewidth}
                 \centering
                \caption{Netflix}
                 \includegraphics[width=0.99\textwidth]{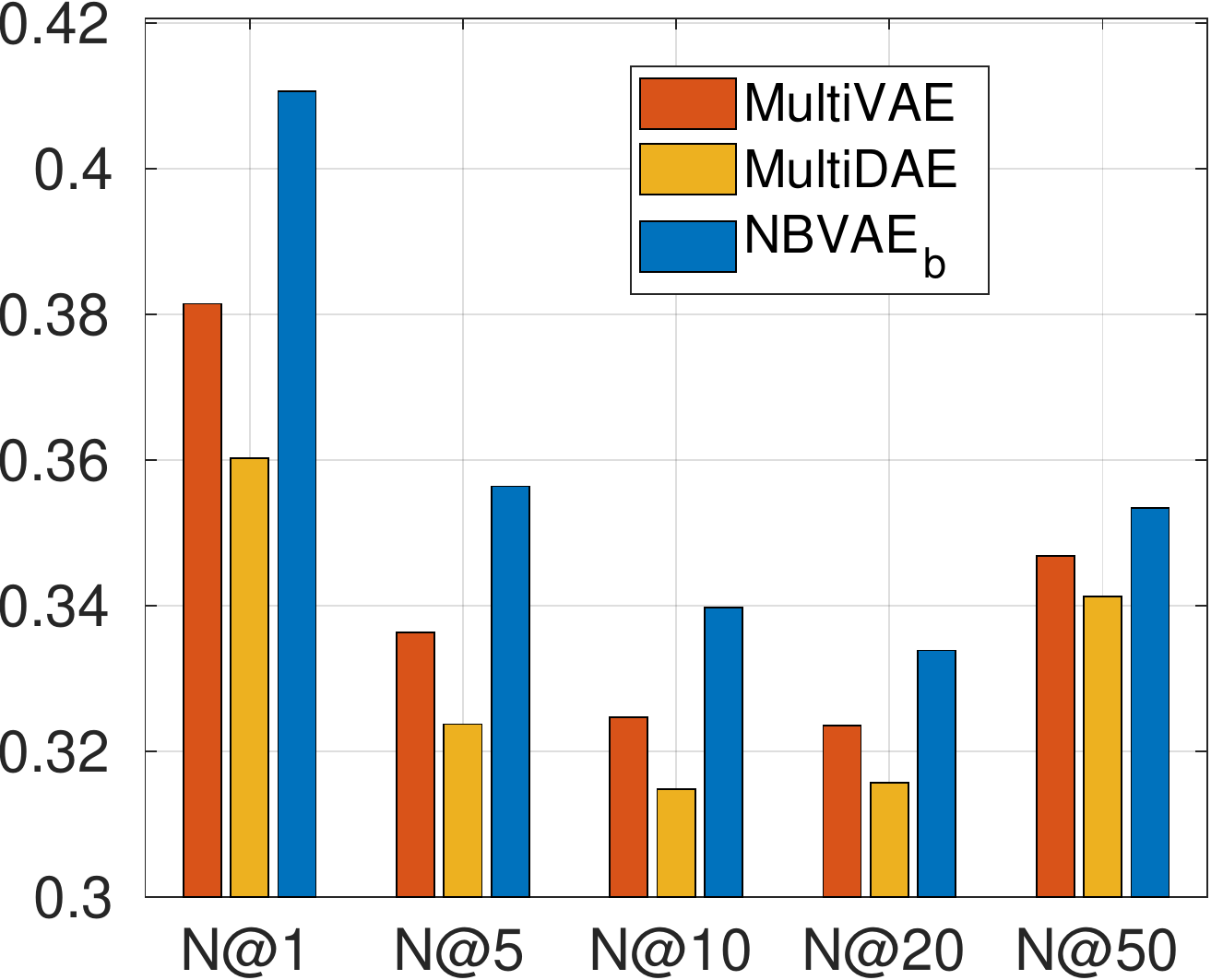}
         \end{subfigure}
         \begin{subfigure}[b]{0.24\linewidth}
                 \centering
                 \caption{MSD}
                 \includegraphics[width=0.99\textwidth]{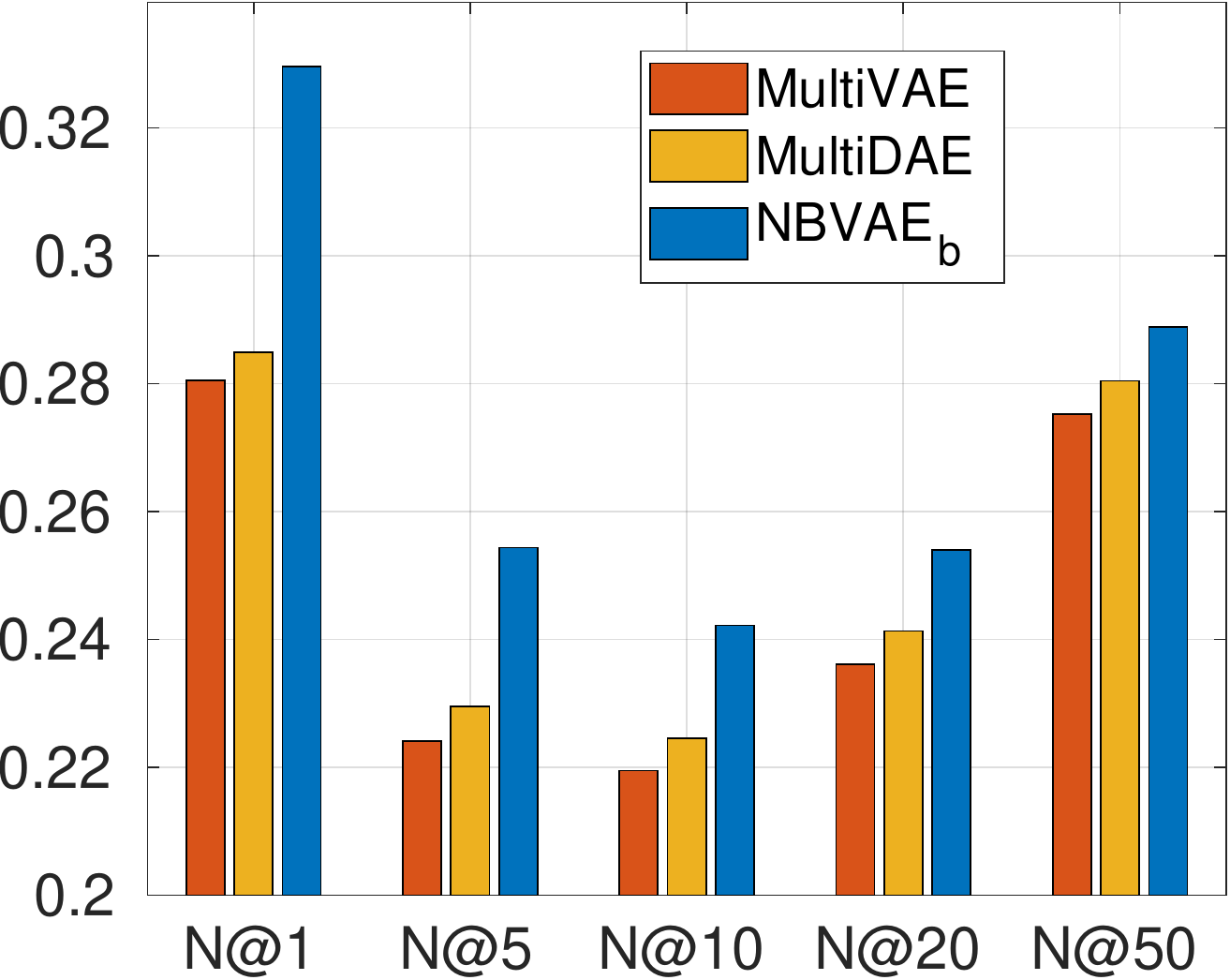}
         \end{subfigure}
         \begin{subfigure}[b]{0.24\linewidth}
                 \centering
                 \includegraphics[width=0.99\textwidth]{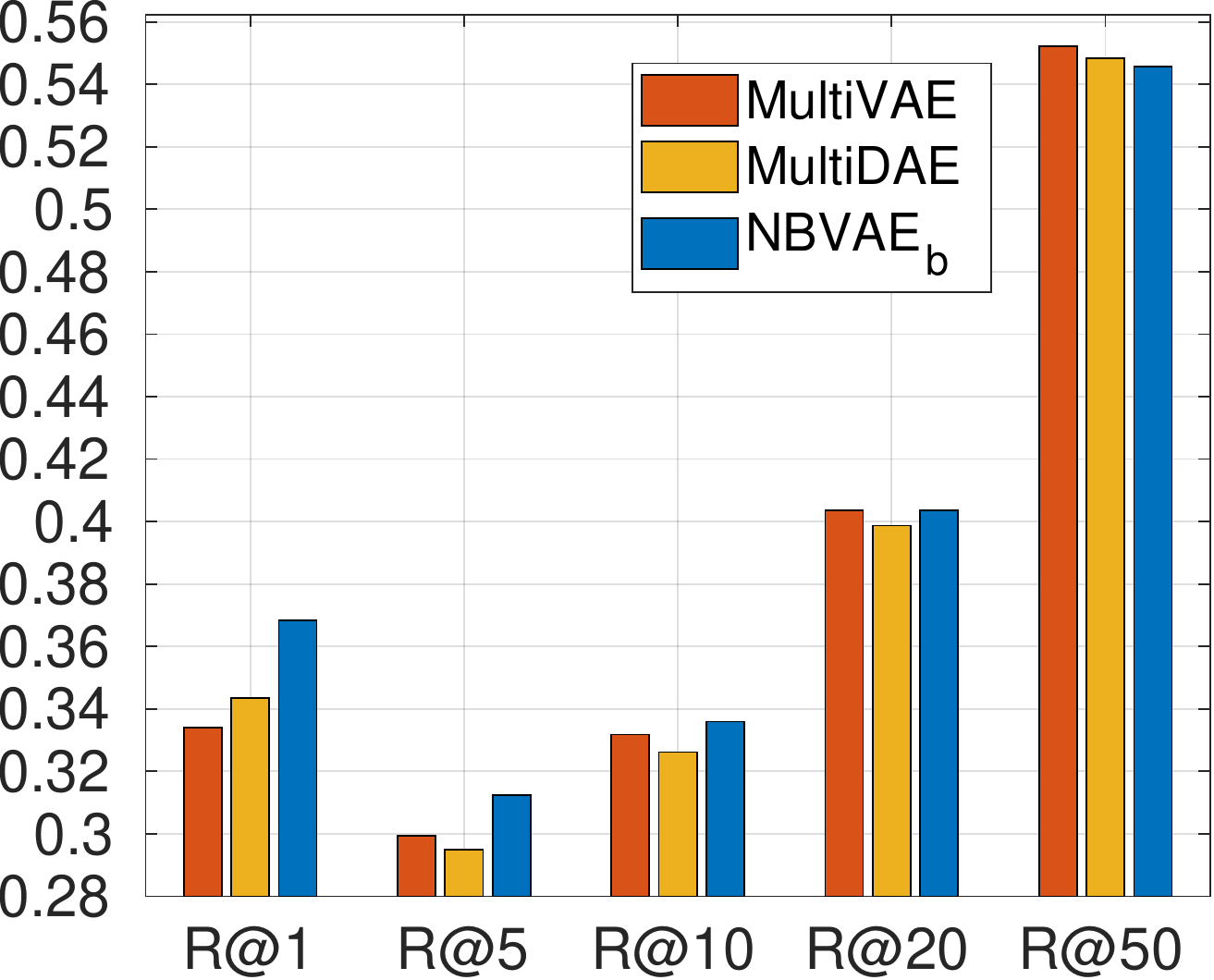}
         \end{subfigure}
         \begin{subfigure}[b]{0.24\linewidth}
                 \centering
                 \includegraphics[width=0.99\textwidth]{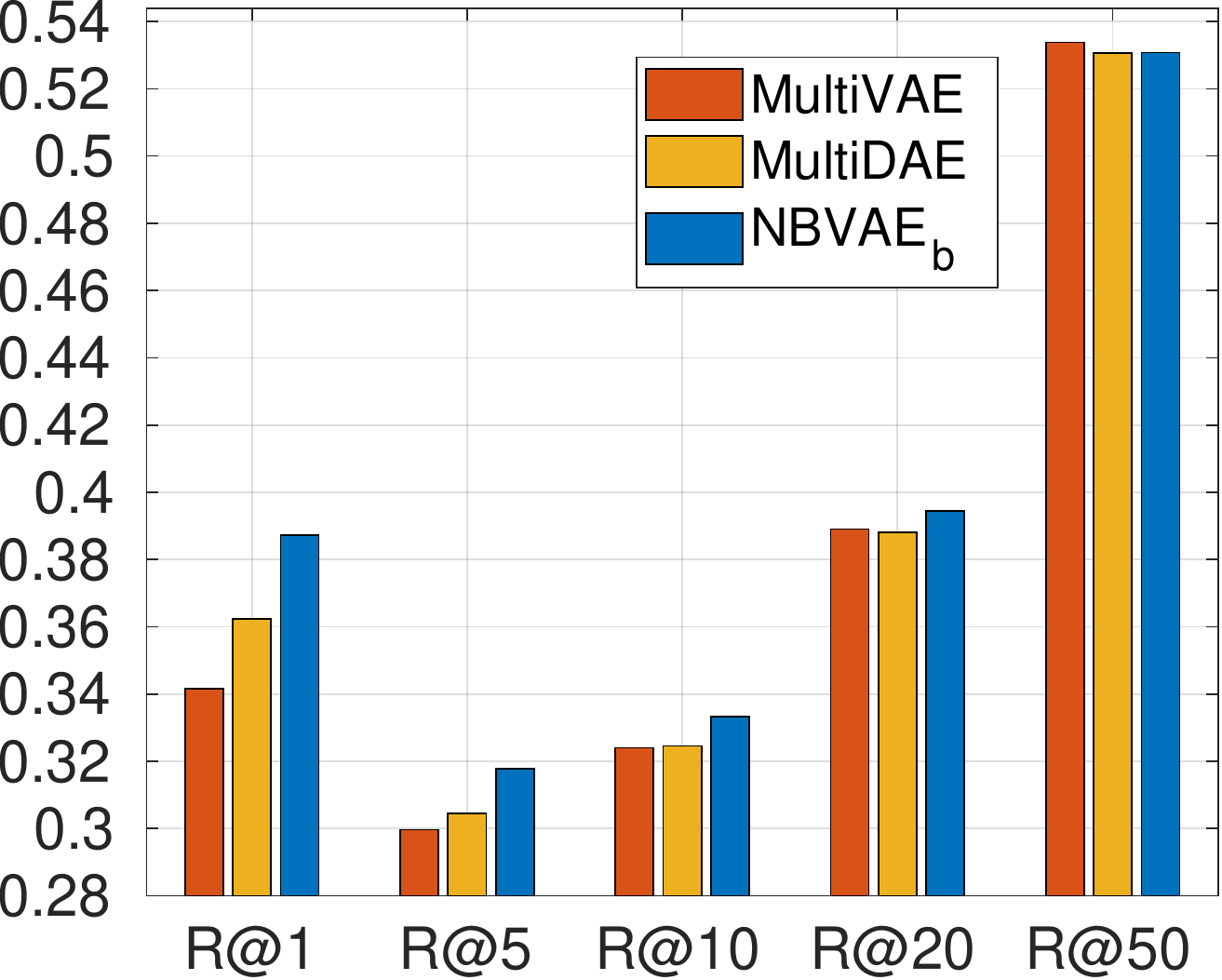}
         \end{subfigure}%
         \begin{subfigure}[b]{0.24\linewidth}
                 \centering
                 \includegraphics[width=0.99\textwidth]{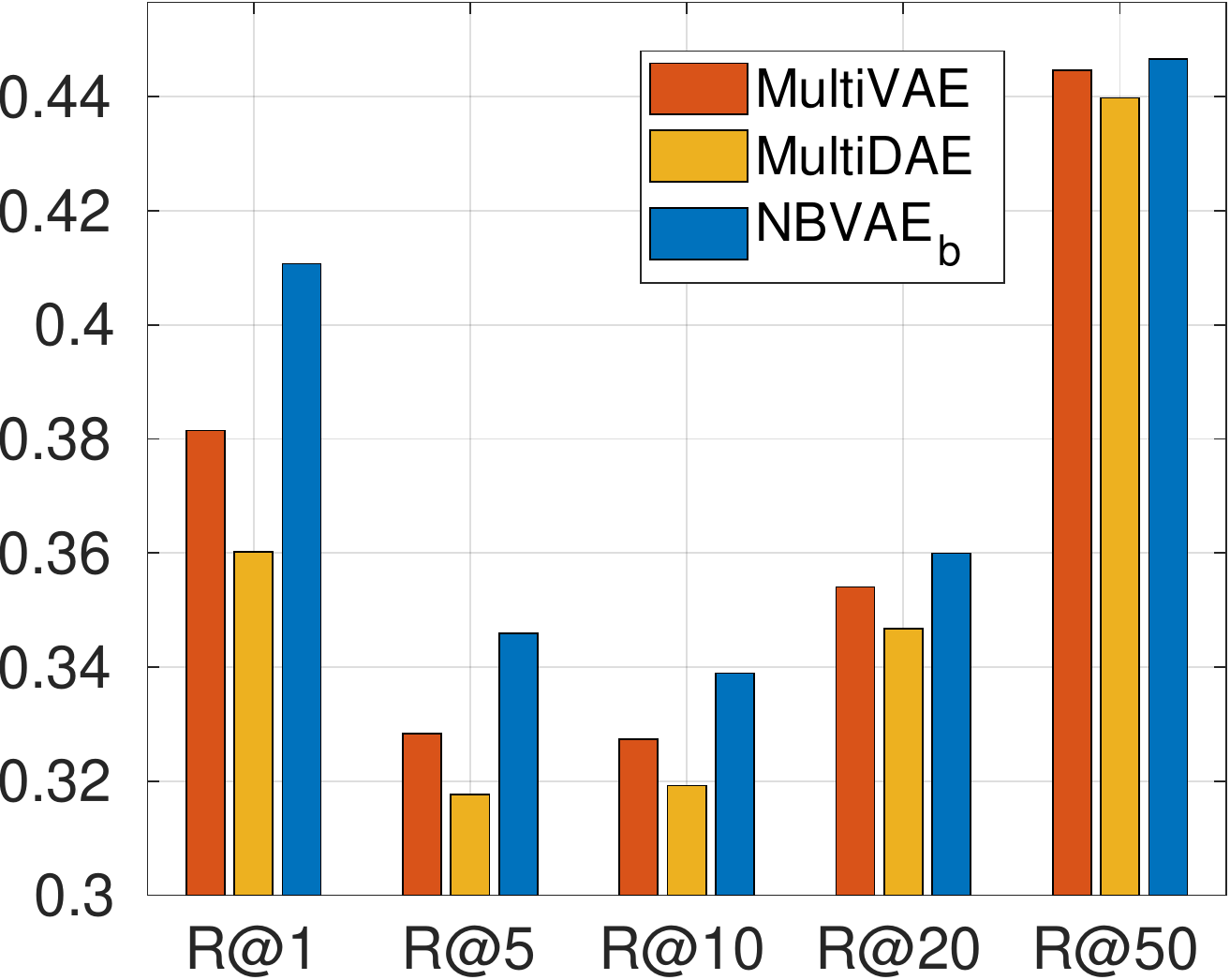}
         \end{subfigure}
         \begin{subfigure}[b]{0.24\linewidth}
                 \centering
                 \includegraphics[width=0.99\textwidth]{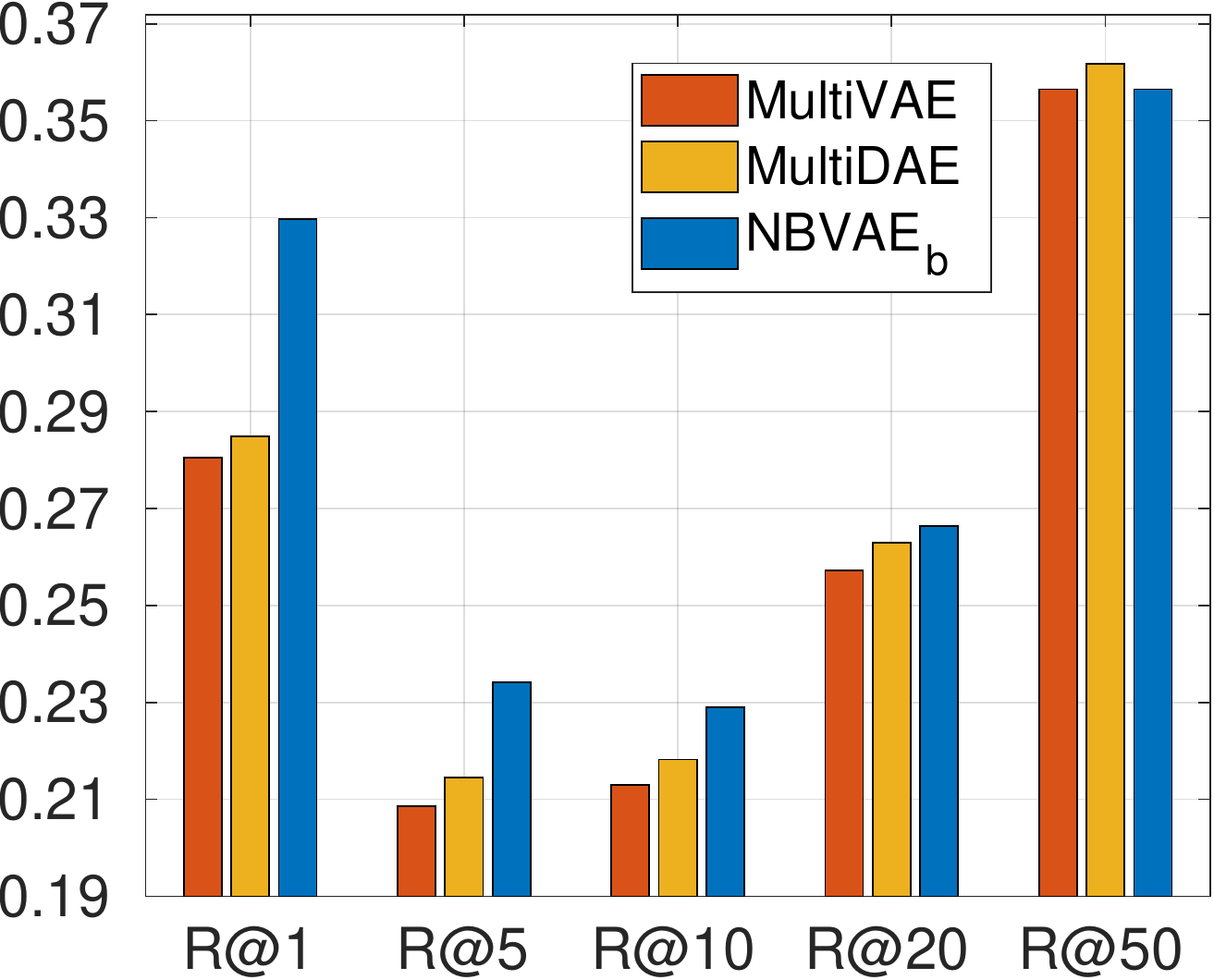}
         \end{subfigure}

\caption{\small Comparisons of NDCG@$R$ (N@$R$) and Recall$R$ (R@$R$). Standard errors in multiple runs are generally less than 0.003 for all the models on all the datasets, which are too tiny to show in the figures.} 
\label{fig-nr}
\vspace{-0.5cm}
\end{figure*}

\subsection{Experiments on collaborative filtering}
Our second set of experiments is targeted at collaborative filtering,
where the task is to recommend items to users using their clicking history.
We evaluate our models' performance on four user-item consumption datasets: MovieLens-10M (ML-10M), MovieLens-20M (ML-20M), Netflix Prize (Netflix), and Million Song Dataset (MSD)~\citep{Bertin-Mahieux2011}.
The details of the datasets are shown in the appendix.
Following~\citet{liang2018variational}, we report two evaluation metrics: Recall@$R$ 
and the truncated normalized discounted cumulative gain (NDCG@$R$). The calculation of the two metrics is shown in the appendix.
As datasets used here are binary, we compared NBVAE$_\text{b}$, 
with the recent VAE models:
\textbf{1)} MultiVAE. \textbf{2)} MultiDAE~\citep{liang2018variational}, a denoising autoencoder (DAE) with multinomial likelihood, which introduces dropout~\citep{srivastava2014dropout} at the input layer. 
MultiVAE and MultiDAE are the state-of-the-art VAE models for collaborative filtering 
and they have been reported to outperform several recent advances such as~\citet{wu2016collaborative} and \citet{he2017neural}.
The experimental settings are consistent with those in~\citet{liang2018variational}, detailed in the appendix.

Figure~\ref{fig-nr} shows the NDCG@$R$ and Recall@$R$ of the models on the four datasets, 
where we used $R \in \{1, 5, 10, 20, 50\}$.
In general, our proposed NBVAE$_{\text{b}}$ outperforms the baselines (i.e.,  
MulitVAE and MultiDAE) on almost all the datasets,
In particular, the margin is notably large while the $R$ value is small, such as 1 or 5.
It indicates that the top-ranked items in NBVAE$_{\text{b}}$ are always more accurate 
than those ranked by MultiVAE and MuliDAE.
This fact is also supported by the large gap of NDCG@$R$ between NBVAE$_{\text{b}}$ and the two baselines, as NDCG@$R$
penalises the true items that are ranked low by the model.
The experimental results show that it is beneficial to treat binary data as binary rather than count-valued. We show more analysis on this point in the appendix.

\subsection{Experiments on multi-label learning}
\label{sec-exp-ml}

Finally, we evaluate the performance of NBVAE$_\text{c}$ on three multi-label learning benchmark datasets: Delicious~\cite{tsoumakas2008effective}, Mediamill~\cite{snoek2006challenge}, and EURLex~\cite{mencia2008efficient}. The details of the datasets and the settings of our model 
 are shown in the appendix. We report Precision@$R$ ($R \in \{1,3,5\}$), which is a widely-used evaluation metric for multi-label learning, following~\citet{jain2017scalable}.
Several recent advances for multi-label learning are treated as the baselines, including LEML~\cite{yu2014large}, PfastreXML~\cite{jain2016extreme}, PD-Sparse~\cite{yen2016pd}, and GenEML~\cite{jain2017scalable}. For the baselines, the reported results are either obtained
using publicly available implementations (with the recommended hyperparameter settings), or the publicly known best results.

\begin{table*}[]
\centering
\caption{\small Precision (P@$R$). Best results for each dataset are in boldface. The standard errors of our model are computed in multiple runs. The results of GenEML on Delicious are not reported due to the unavailability.}
\label{tb-ml}
\resizebox{0.7\textwidth}{!}{
\begin{tabular}{|c|c|c|c|c|c|c|}
\hline
Datasets  & Metric                                                  & LEML                                                          & PfastreXML                                                    & PD-Sparse                                                     & GenEML                                                        & NBVAE$_\text{c}$                                                        \\ \hline
Delicious & \begin{tabular}[c]{@{}c@{}}P@1\\ P@3\\ P@5\end{tabular} & \begin{tabular}[c]{@{}c@{}}65.67\\ 60.55\\ 56.08\end{tabular} & \begin{tabular}[c]{@{}c@{}}67.13\\ \textbf{63.48}\\ \textbf{60.74}\end{tabular} & \begin{tabular}[c]{@{}c@{}}51.82\\ 46.00\\ 42.02\end{tabular} & \begin{tabular}[c]{@{}c@{}}-\\ -\\ -\end{tabular}             & \begin{tabular}[c]{@{}c@{}}\textbf{68.49}$\pm$0.39\\ 62.83$\pm$0.47\\ 58.04$\pm$0.31\end{tabular} \\ \hline
Mediamill & \begin{tabular}[c]{@{}c@{}}P@1\\ P@3\\ P@5\end{tabular} & \begin{tabular}[c]{@{}c@{}}84.01\\ 67.20\\ 52.80\end{tabular} & \begin{tabular}[c]{@{}c@{}}83.98\\ 67.37\\ 53.02\end{tabular} & \begin{tabular}[c]{@{}c@{}}81.86\\ 62.52\\ 45.11\end{tabular} & \begin{tabular}[c]{@{}c@{}}87.15\\ 69.9\\ 55.21\end{tabular}  & \begin{tabular}[c]{@{}c@{}}\textbf{88.27}$\pm$0.24\\ \textbf{71.47}$\pm$0.18\\ \textbf{56.76}$\pm$0.26\end{tabular} \\ \hline
EURLex & \begin{tabular}[c]{@{}c@{}}P@1\\ P@3\\ P@5\end{tabular} & \begin{tabular}[c]{@{}c@{}}63.40\\ 50.35\\ 41.28\end{tabular} & \begin{tabular}[c]{@{}c@{}}75.45\\ 62.70\\ 52.51\end{tabular} & \begin{tabular}[c]{@{}c@{}}76.43\\ 60.37\\ 49.72\end{tabular} & \begin{tabular}[c]{@{}c@{}}77.75\\ 63.98\\ 53.24\end{tabular} & \begin{tabular}[c]{@{}c@{}}\textbf{78.28}$\pm$0.49\\ \textbf{66.09}$\pm$0.17\\ \textbf{55.47}$\pm$0.15\end{tabular} \\ \hline
\end{tabular}
}
\vspace{-0.5cm}
\end{table*}

Table~\ref{tb-ml} shows the performance comparisons on the multi-label learning datasets. It can be observed that the proposed NBVAE$_\text{c}$ generally performs the best than the others on the prediction precision, showing its promising potential on multi-label learning problems. Note that the baselines are specialised to the multi-label learning problem, many of which take multiple steps of processing of the labels and features or use complex optimisation algorithms. Compared with those models, the model simplicity of NBVAE$_\text{c}$ is another appealing advantage.

\section{Conclusion}

In this paper, we have focused on analysing and addressing two issues of PMF models on large-scale, sparse, discrete data: insufficient capability of modelling overdispersion in count-valued data and model misspecification in binary data.
We have tackled those two issues by developing a VAE-based framework named NBVAE, which generates discrete data from the negative-binomial distribution. Specifically, the predictive distribution of NBVAE shows that the model can explicitly capture self-excitation because of the use of NB, and its deep structure offers better model capacity on capturing cross-excitation. By better modelling the two kinds of excitations, NBVAE is able to better handle overdispersion in count-valued data.
In addition, we have developed two variants of NBVAE, NBVAE$_\text{b}$ and NBVAE$_\text{c}$,  which are able to achieve better modelling performance on binary data and on multi-label learning. Extensive experiments have shown that NBVAE, NBVAE$_\text{b}$, and NBVAE$_\text{c}$ are able to achieve the state-of-the-art performance on text analysis, collaborative filtering, and multi-label learning.

\bibliographystyle{plainnat}
\bibliography{nbvae}



\section*{Supplementary material}

# Supplementary Material for "Variational Autoencoders for Sparse and Overdispersed Discrete Data"

He Zhao*     Piyush Rai†     Lan Du*     Wray Buntine*

Mingyuan Zhou‡

*Monash University, Australia
†Indian Institute of Technology, Kanpur, India
‡The University of Texas at Austin, USA

## 1 Experiments on text analysis

### 1.1 Datasets

The statistics of the datasets used in the text analysis experiments are shown in Table 1. The 20NG and RCV datasets were downloaded from the code repository of Gan et al. [3][1]. The Wiki dataset was downloaded from *Wikipedia* using the scripts provided in Hoffman et al. [5].

### 1.2 Evaluation metric

We report per-heldout-word perplexity of all the models, which is a widely-used metric for text analysis. Following the approach in Wallach et al. [12], after training a model with the training documents, we randomly selected some words as the observed words and used the remaining words as the unobserved words in each testing document, then used the observed words to estimate the predictive probability, and finally computed the perplexity of the unobserved words. Specifically, suppose that the matrix of the testing documents is $\mathbf{Y}^* \in \mathbb{N}^{V \times N_{\text{test}}}$, which is split into the observed word matrix $\mathbf{Y}^{*\text{o}} \in \mathbb{N}^{V \times N_{\text{test}}}$ and the unobserved word matrix $\mathbf{Y}^{*\text{u}} \in \mathbb{N}^{V \times N_{\text{test}}}$, where $\mathbf{Y}^* = \mathbf{Y}^{*\text{o}} + \mathbf{Y}^{*\text{u}}$. The predictive rates of the testing documents are estimated with $\mathbf{Y}^{*\text{o}}$ and used to compute the perplexity of $\mathbf{Y}^{*\text{u}}$, detailed as follows[2]:

$$\text{Perplexity} = \exp - \left( \frac{1}{y_{\cdot\cdot}^{*\text{u}}} \sum_{j}^{N_{\text{test}}} \sum_{v}^{V} y_{vj}^{*\text{u}} \log \frac{l_{vj}}{l_{\cdot j}} \right), \tag{1}$$

where $y_{\cdot\cdot}^{*\text{u}} = \sum_{j}^{N_{\text{test}}} \sum_{v}^{V} y_{vj}^{*\text{u}}$. Note that $l_{vj}$ is the predictive rate, whose derivation is model specific shown in Table 1 of the main paper. To conduct a fair comparison of perplexity, the settings of the datasets including the preprocessing process, the training/testing splits, the observed/unobserved words splits, and the perplexity calculation are consistent with those in Gan et al. [3], Henao et al. [4], Cong et al. [2].

---

[1] `https://github.com/zhegan27/dpfa_icml2015`

[2] Our perplexity calculation is the same with the ones in Gan et al. [3], Henao et al. [4], Cong et al. [2], but different from the ones in Miao et al. [11, 10], Krishnan et al. [8], which use ELBO obtained from all the words of a testing document without splitting it. The results of Miao et al. [11, 10], Krishnan et al. [8] can only be compared with models with variational inference.

Table 1: Statistics of the datasets in text analysis. $N_{\text{train}}$: number of training instances, $N_{\text{test}}$: number of test instances. The number of nonzeros and density are computed of each whole dataset.

| Dataset | $N_{\text{train}}$ | $N_{\text{test}}$ | V | #Nonzeros | Density |
|---|---|---|---|---|---|
| 20NG | 11,315 | 7,531 | 2,000 | 774,984 | 0.0343 |
| RCV | 794,414 | 10,000 | 10,000 | 58,637,816 | 0.0074 |
| Wiki | 10,000,000 | 1,000 | 7,702 | 82,311,745 | 0.0107 |

Table 2: Perplexity comparisons with larger layer width

| Model | Inference | Layers | RCV | Wiki |
|---|---|---|---|---|
| DLDA | TLASGR | 256-128-64 | 710 | 682 |
| NBFA | Gibbs | 256 | 649 | - |
| MultiVAE | VAE | 256-128 | 587 | 589 |
| NBVAE$_{\text{dm}}$ | VAE | 256-128 | 552 | 462 |
| NBVAE | VAE | 256-128 | 535 | 451 |
| DLDA | TLASGR | 512-256-128 | 656 | 602 |
| MultiVAE | VAE | 512-256 | 552 | 558 |
| NBVAE$_{\text{dm}}$ | VAE | 512-256 | 517 | 451 |
| NBVAE | VAE | 512-256 | 512 | 445 |

### 1.3 Experimental settings

In the experiments of text analysis, in terms of model settings of our proposed models, following [9], we basically used the settings as for MultiVAE. Specifically, for both MultiVAE, NBVAE, and NBVAE$_{\text{dm}}$,

- We applied the fully connected multi-layer perceptrons (MLP) with tanh as the nonlinear activation function between the layers of the encoder and the decoder.
- We used the same network architecture for the two parametric functions in the decoder, $f_{\theta^r}(\cdot)$ and $f_{\theta^p}(\cdot)$.
- The architecture of $f_{\phi}(\cdot)$ is symmetric to those of $f_{\theta^r}(\cdot)$ and $f_{\theta^p}(\cdot)$. For example, if we use [32, 64, 128] as the architecture of the hidden layers for the decoder, then $K = 32$ is the dimension of the latent representations and the architecture of the hidden layers for the encoder would be [128, 64, 32].
- The output layers of the encoder and decoder have no activation function.
- We set the batch size to 500 and 2000 for 20NG and the other two larger datasets, respectively.
- The number of training epochs was set to 800 and the optimisation of the VAE models was done by Adam [7] with 0.003 as the learning rate.
- We used the same KL annealing procedure mentioned in the MultiVAE paper [9].

For the baselines, we used the original model settings provided in the code published by the authors. For the VAE-based models, we report the perplexity computed with the parameters (the encoder and decoder) in the last iteration of the training phrase, whereas for models with MCMC sampling (e.g., NBFA), we report the perplexity averaged over multiple samples in the collection iterations.

### 1.4 In-depth experiments

#### 1.4.1 Perplexity with larger layer width

In Table 3 of the main paper, we followed the settings of Gan et al. [3], Henao et al. [4], Cong et al. [2] to set up the network structure, where 128 was used as the maximum layer width. Given the sizes of RCV and Wiki, we increased the maximum to 256 and 512 to further compare the results of related models and study if the width of layer matters in the modelling. The results are shown in Table 2. If jointly looking at Table 3 of the main paper and Table 2, we find:

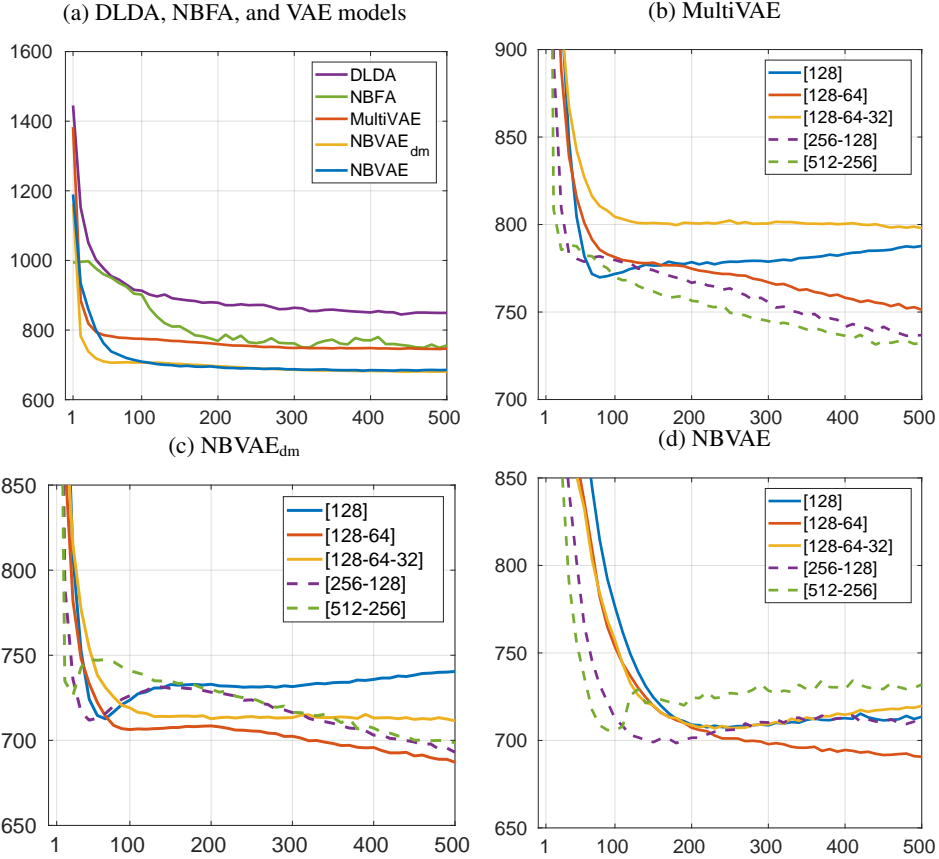

Figure 1: Perplexity of the testing documents over training iterations on the 20NG dataset. (a) Perplexity of DLDA ([128-64-32]), NBVAE ([128]), and VAE models ([128-64]). Unlike in Table 3 of the main paper, where we report the averaged perplexity for DLDA and NBFA, here we report their point perplexity with a single sample, which is expected to be higher than the averaged perplexity. (b)-(d) Perplexity of MultiVAE, NBVAE$_{dm}$, and NBVAE, respectively. Here we varied the depth of the layers and the length of each layer.

- With the increased layer widths, all the models have gained significant improvements.
- Among all the models, regardless of each layer's width, both NBVAE and NBVAE$_{dm}$ outperform the others with a significant margin.
- The table also shows that with smaller model structures (e.g., [256-128]), NBVAE and NBVAE$_{dm}$ is able to achieve comparable results with the other models with larger structures, further demonstrating our models' expressiveness.

### 1.4.2 Comparisons of convergence speed

Here we compare the perplexity convergence speed of our models with DLDA, NBFA, and MultiVAE, in Figure 1a. It can be seen that VAE models enjoys faster convergence speed than DLDA and NBFA using MCMC sampling. Moreover, to estimate the posterior of DLDA and NBFA, we need to average over multiple samples. To do so, the algorithm needs to iterate over the testing documents multiple times. Whereas our models are able to derive good results with a single sample.

### 1.4.3 Overfitting/Underfitting of VAE models

Unlike conventional Bayesian models with hierarchical prior distributions such as DLDA, it is known that VAE models are sensitive to the dimensions of the latent representations and the encoder/decoder structures. Here we study how the performance of the VAE models on 20NG varies with different settings, shown in Figure 1b to 1d.

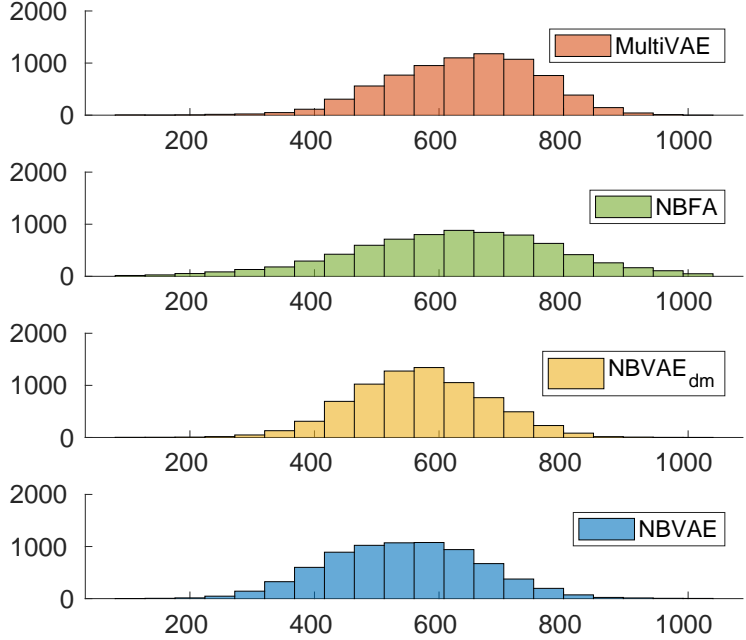

Figure 2: Comparisons of the entropy histograms on the 20NG dataset.

It can be seen that increasing the layer widths from [128-64] to [256-128] and [512-256] improves MultiVAE's performance but it is not the case for NBVAE and NBVAE$_{dm}$. This is because that NBVAE and NBVAE$_{dm}$ have more parameters than MultiVAE and increasing the parameter space will be more likely to cause overfitting in the two models. As NBVAE is with more parameters than NBVAE$_{dm}$, the overfitting in NBVAE is more obvious[3].

Now we compare the cases where the encoder/decoder structures go from [128] to [128-64] and [128-64-32]. We can observe a similar trend in the three models, i.e., [128-64] performs the best and going deeper or shallower gets inferior performance. Using the settings of [128] means that we have a large dimension of the latent representations but the encoder/decoder networks are quite shallow, which may not be able to learn good representations. On the contrary, with [128-64-32], although the networks are deep, the dimensions of the latent representations may be too small to learn a good model. This set of experiments demonstrate that model selection is important to the VAE models that we are comparing with, but studying the details is outside the paper and we leave it for future study.

**Comparisons of predictive distribution entropy** To further analyse the predictive distributions of related models, we compare the entropy of the predictive distributions on the testing documents of 20NG. The entropy of document $j$ given the predictive distribution parameterised with $l'_j$ is computed as follows:

$$\text{entropy}_j = \exp\left(-\sum_v^V l'_{vj} \log(l'_{vj})\right), \quad (2)$$

where $l'_j$ is assumed to be normalised and is computed model-specifically according to Table 1 of the main paper.

After computing the entropy of all the testing documents, we plot the histograms in Figure 2. The intuition here is that the entropy of the predictive distribution of a document can be viewed as the effective number of unique words that the document is expected to focus on, given a model. Therefore, if a model takes self-excitation into account, the entropy of a document is expected to be small, because the model's predictive distribution will put large mass on the words that already occur in the document. That is to say, the document is expected to focus on the existing words. Whereas if a model without the consideration of self-excitation, its predictive distribution of a document would

---
[3]Note that compared with RCV and Wiki, 20NG is a relatively small dataset. Therefore, unlike results in Table 2, using wider layers is easier to cause overfitting.

4

Table 3: Statistics of the datasets in collaborative filtering. $N_{\text{train}}$: number of training instances, $N_{\text{test}}$: number of test instances. The number of nonzeros and density are computed of each whole dataset.

| Dataset | $N_{\text{train}}$ | $N_{\text{test}}$ | V | #Nonzeros | Density |
|---|---|---|---|---|---|
| ML-10M | 49,167 | 10,000 | 10,066 | 4,131,372 | 0.0059 |
| ML-20M | 116,677 | 10,000 | 20,108 | 9,128,733 | 0.0033 |
| Netflix | 383,435 | 40,000 | 17,769 | 50,980,816 | 0.0062 |
| MSD | 459,330 | 50,000 | 36,716 | 29,138,887 | 0.0014 |

relatively spread over all the words. Figure 2 shows that the average entropy of our models is smaller than that of MultiVAE, demonstrating the effect of modelling self-excitation.

## 2 Experiments on collaborative filtering

### 2.1 Datasets

MovieLens-10M (ML-10M) and MovieLens-20M (ML-20M) were downloaded from `https://grouplens.org/datasets/movielens/`; Netflix Prize (Netflix) was downloaded from `http://www.netflixprize.com/`; Million Song Dataset (MSD) was downloaded from `https://labrosa.ee.columbia.edu/millionsong/` [1]. All the datasets were preprocessed and binarised by the Python code provided by [9], using the same settings described in the paper. The statistics of the datasets are shown in Table 1. Note that following [9], we also generated a validation set with the same size of the testing set.

### 2.2 Evaluation metrics

Two ranking-based metrics were used. They are Recall@$R$ and the truncated normalized discounted cumulative gain (NDCG@$R$). To compute those metrics, following [9], we first estimated the predictive rate $l'_j$ of user $j$ given the observed items $y_j^{*o}$, and then ranked the unobserved items $y_j^{*u}$ by sorting $l'_j$. The metrics are computed as follows:

$$\text{Recall@}R = \frac{\sum_{r=1}^{R} \mathbf{1}\left(y_{\omega(r)j}^{*u} = 1\right)}{\min(R, y_{\cdot j}^{*u})}, \quad (3)$$

$$\text{DCG@}R = \frac{\sum_{r=1}^{R} 2^{\mathbf{1}\left(y_{\omega(r)j}^{*u} = 1\right)} - 1}{\log(r+1)}, \quad (4)$$

where $\omega(r) \in \{1, \cdots, V\}$ is the item at rank $r$, obtained by sorting the predictive rate of the user; $\mathbf{1}\left(y_{\omega(r)j}^{*u} = 1\right)$ indicates whether the item is actually clicked on by user $j$; NDCG@$R$ is computed by linearly normalising DCG@$R$ into $[0, 1]$. Intuitively, Recall@$R$ measures the number of the $R$ predicted items that are within the set of the ground-truth items but does not consider the item rank in $R$, while NDCG@$R$ assigns larger discounts to lower ranked items. In the experiments, we used the code provided in [9] to compute the above two metrics. Moreover, we report the testing performance of the models with the best NDCG@50 on the validation set.

### 2.3 Experimental settings

For NBVAE, NBVAE$_{\text{dm}}$, and NBVAE$_{\text{b}}$, we used the same settings as in the text analysis experiments, except that:

- The batch size was set to 500 for all the datasets.
- We used two hidden layers in the encoder with [200-600] (The architectures of the two parametric functions in the decoder are symmetric to that in the encoder).
- Following MultiVAE, we used the annealing cap $\beta$, which was set to 0.2, detailed in Liang et al. [9].

Note that all the above settings are consistent with those in [9] The original code of MultiVAE and MultiDAE and their best settings provided by the authors were used in the comparison.

Table 4: NDCG@$R$ (N@$R$) and Recall$R$ (R@$R$) of NBVAE and its variants on ML-10M. Best results are in boldface.

| Model | N@1 | N@5 | N@10 | N@20 | N@50 | R@1 | R@5 | R@10 | R@20 | R@50 |
|---|---|---|---|---|---|---|---|---|---|---|
| NBVAE | 0.3333 | 0.2951 | 0.3012 | 0.3263 | 0.3788 | 0.3333 | 0.2927 | 0.3224 | 0.3968 | 0.5453 |
| NBVAE$_{dm}$ | 0.3441 | 0.3034 | 0.3070 | 0.3294 | 0.3800 | 0.3441 | 0.3005 | 0.3270 | 0.3978 | 0.5441 |
| NBVAE$_b$ | **0.3684** | **0.3187** | **0.3198** | **0.3394** | **0.3878** | **0.3684** | **0.3124** | **0.3360** | **0.4039** | **0.5456** |

Table 5: The statistics of the datasets used in the experiments. $N_{\text{train}}$: number of training instances, $N_{\text{test}}$: number of test instances, $D$: number of features, $V$: number of labels.

| Dataset | $N_{\text{train}}$ | $N_{\text{test}}$ | $D$ | $V$ |
|---|---|---|---|---|
| Delicious | 12920 | 3185 | 500 | 983 |
| Mediamill | 30993 | 12914 | 120 | 101 |
| EURLex | 15539 | 3809 | 5000 | 3993 |

### 2.4 In-depth experiments

To further demonstrate the benefit of using NBVAE$_b$, we compare it with NBVAE and NBVAE$_{dm}$ on the ML-10M dataset, where the latter two models treat binary data as count-valued data. The results of NDCG@$R$ and Recall@$R$ on ML-10M are shown in Table 4. It can be observed that NBVAE$_b$'s results are significantly better than NBVAE and NBVAE$_{dm}$, showing the necessity of dealing with binary data separately from count-valued data.

## 3 Experiments on Multi-Label Learning

### 3.1 Datasets

All the datasets were downloaded from `http://manikvarma.org/downloads/XC/XMLRepository.html` and the statistics of the datasets are shown in Table 5.

### 3.2 Evaluation metrics

We report Precision@$R$ ($R \in \{1, 3, 5\}$), which is a widely-used ranking-based evaluation metric for multi-label learning, following Jain et al. [6]. To compute this metric, after training NBVAE$_c$, given the feature vector of a testing sample $j^*$, we can feed $\boldsymbol{x}_{j^*}$ into the feature encoder to sample the latent representation, $\boldsymbol{z}_{j^*}$, then feed it into the decoder to get the predictive rate $\boldsymbol{l'}_{j^*}$. With the predictive rate, we can rank the labels and compute Precision@$R$, which is similar to the computation of Recall and NDCG used in collaborative filtering.

### 3.3 Experimental settings

For NBVAE$_c$ in multi-label learning, we used the same settings as NBVAE$_b$ in the text analysis experiments, specifically:

- In the Delicious and Mediamill datasets, we used [200-600] for two hidden layers in the encode and for EURLex, we used one hidden layer in the encoder with 600 units.
- We used relu as the activation function for Delicious and EURLex and tanh as the activation function for Mediamill.

### 3.4 In-depth experiments

In this section, we empirically demonstrate our proposed inference scheme of NBVAE$_c$ detailed in Section 3 of the main paper. Specifically, we ran the model with the following three settings: (a) In each and every training iteration, we always sample the latent representations from the encoder, i.e., $\boldsymbol{z}_j \sim \mathrm{q}(\boldsymbol{z}_j \mid \boldsymbol{y}_j)$; (b) In each and every training iteration, we always sample the latent representations from the feature encoder, i.e., $\boldsymbol{z}_j \sim \mathrm{q}(\boldsymbol{z}_j \mid \boldsymbol{x}_j)$; (c) We alternatively sample the latent representations from the encoder and the feature encoder, e.g., in one iteration, $\boldsymbol{z}_j \sim \mathrm{q}(\boldsymbol{z}_j \mid \boldsymbol{y}_j)$ and in the next

Table 6: Precision results on the Delicious dataset for NBVAE$_c$.

| Precision | (a) $z_j \sim q(z_j \mid y_j)$ | (b) $z_j \sim p(z_j \mid x_j)$ | (c) Alternative |
|---|---|---|---|
| P@1 | 65.06±0.35 | 67.26±0.27 | **68.49**±0.39 |
| P@3 | 59.88±0.30 | 61.74±0.10 | **62.83**±0.47 |
| P@5 | 54.92±0.15 | 56.86±0.20 | **58.04**±0.31 |

iteration, $z_j \sim q(z_j \mid x_j)$. The last one is the proposed inference scheme. Figure 6 shows the results of NBVAE$_c$ with the three settings on the Delicious dataset.

It can be shown that the proposed scheme (c) performs significantly better than the others, demonstrating its effectiveness in multi-label learning.



\end{document}